\documentclass[a4paper]{IEEEtran}
\IEEEoverridecommandlockouts
\usepackage{cite}
\usepackage[hidelinks]{hyperref}
\usepackage{amsmath,amssymb,amsfonts}
\usepackage{algorithmic}
\usepackage{graphicx}
\usepackage{textcomp}
\usepackage{xcolor}
\usepackage{physics}
\usepackage{multirow}
\usepackage{amsmath, bm}
\usepackage{booktabs}
\usepackage{caption}
\usepackage[font=footnotesize]{subcaption}
\graphicspath{ {./images/} }
\def\BibTeX{{\rm B\kern-.05em{\sc i\kern-.025em b}\kern-.08em
    T\kern-.1667em\lower.7ex\hbox{E}\kern-.125emX}}

\newcommand{\nhat}[1]{#1}
\newcommand{\trung}[1]{#1}
\newcommand{\hoang}[1]{#1}
\newcommand{\tung}[1]{#1}
\begin{document}

\title{Variational Autoencoder for Anomaly Detection: \\ A Comparative Study\\
}

\author{\IEEEauthorblockN{\textit{Huy Hoang Nguyen $^ *$, Cuong Nhat Nguyen $^ *$, Xuan Tung Dao$^ *$, Quoc Trung Duong$^ *$, Dzung Pham Thi Kim$^ *$, \textit{Minh-Tan Pham$^{\dagger}$}}}
\\
\IEEEauthorblockA{$^ *$FPT University, Swinburne Vietnam, Hanoi, Vietnam \\
}
\IEEEauthorblockA{$^{\dagger}$IRISA, Université Bretagne Sud, UMR 6074, 56000 Vannes, France \\
}

}




\maketitle

\begin{abstract}
\trung{
This paper aims to conduct a comparative analysis of contemporary Variational Autoencoder (VAE) architectures employed in anomaly detection, elucidating their performance and behavioral characteristics within this specific task. The architectural configurations under consideration encompass the original VAE baseline, the VAE with a Gaussian Random Field prior (VAE-GRF), and the VAE incorporating a vision transformer (ViT-VAE). The findings reveal that ViT-VAE exhibits exemplary performance across various scenarios, whereas VAE-GRF may necessitate more intricate hyperparameter tuning to attain its optimal performance state. Additionally, to mitigate the propensity for over-reliance on results derived from the widely used MVTec dataset, this paper leverages the recently-public MiAD dataset for benchmarking. This deliberate inclusion seeks to enhance result competitiveness by alleviating the impact of domain-specific models tailored exclusively for MVTec, thereby contributing to a more robust evaluation framework. Codes is available at \href{https://github.com/endtheme123/VAE-compare.git}{this link}.}
\end{abstract}


\section{Introduction}
    
\hoang{Anomaly detection (AD) represents a machine learning process designed to discern abnormal patterns within a given set of input data. This process finds application across diverse fields, including but not limited to fraudulent detection within the banking industry, intrusion detection in security systems, anomaly detection in manufacturing, and cancer detection within healthcare \cite{nassif2021machine}. There are also various ways to implement an anomaly detection model ranging from supervised anomaly detection \cite{li2023target}, to semi-supervised anomaly detection \cite{akcay2018ganomaly}, and unsupervised anomaly detection \cite{gangloff2023variational}. In this paper, we will focus on unsupervised anomaly detection.

Reconstruction-based AD constitutes a specific branch of anomaly detection that identifies abnormal patterns through the reconstruction capacity of deep neural networks. This variant aims to reconstruct input data using generative deep models and subsequently compares the reconstructed data with the original input to identify and localize anomalies. While there exist numerous reconstruction-based networks suitable for the AD task, our focus in this work is on Variational Autoencoder (VAE) and its various adaptations.

In recent years, generative architectures such as Generative Adversarial Networks (GANs), Diffusion models, and VAEs have garnered significant attention due to their extensive applications across various facets of human society. Despite being the oldest among these architectures, VAE has continued to evolve and maintain its prominence in the realm of anomaly detection. Gangloff et al. \cite{gangloff2023unsupervised} previously introduced a VAE architecture with Gaussian Random Field priors (VAE-GRF) and showed that this model exhibited competitive results compared to the standard VAE when applied to textural images. The same authors have investigated the discrete latent space of Vector-Quantized VAEs and introduced a novel alignment map to measure anomaly scores for AD from images \cite{gangloff2022leveraging,pham2023weakly}.

From the literature, there has been several research studies showing interests in integrating attention mechanisms \cite{vaswani2017attention} into the encoder-decoder AE architecture. In \cite{zhang2023exploring, choi2022viv, lee2022anovit}, the implementation of attention for these systems is based on the Vision Transformer (ViT) \cite{dosovitskiy2020image} idea for feature extraction, followed by the utilization of convolutional layers to reconstruct the image directly from image feature instead of the traditional ViT class tokens. Additional research works have explored Masked Autoencoder for anomaly detection \cite{huang2022self,georgescu2023masked} by using the masking trick of the Transformer for ViT's image feature, which involves masking part of the image patches for self-supervised learning. However, all the aforementioned research has used Transformer in the AE framework and not the VAE. For instant, \cite{chen2023learning} has been found to be among limited studies from which ViT is implemented in a VAE architecture. Nevertheless, instead of using image features and Convolutional layers to reconstruct the image, it employs class tokens and Linear layers respectively with multiple pre-processing steps to enhance the input information. Thus, one of the goal of this work is to implement a VAE using ViT's image features to perform AD task for comparative purposes.

In summary, the main contributions in this paper are:
\begin{itemize}
    \item First, we investigate and implement a ViT-VAE framework for anomaly detection and elucidate its effectiveness in this context. Inspired from \cite{gangloff2023unsupervised, gangloff2023variational}, we train ViT-VAE in an unsupervised manner and measure anomaly scores from both image space and the latent space.
    \item Second, we conduct extensive experiments to evaluate and benchmark the performance of contemporary VAE architectures including the standard VAE, the VAE-GRF and the ViT-VAE on two public datasets including the MVTec AD \cite{bergmann2019mvtec} and the MiAD \cite{bao2023miad}.
\end{itemize}

}

\section{Methodology}
\label{sec:method}
\subsection{VAE Architectures}


\nhat{Anomaly detection identifies unusual patterns in data, which is challenging for traditional methods \cite{munir2019comparative}. As one of the most popular deep learning-based generative models, VAEs address this task by learning a latent space representation with both mean and variability in data distribution, making them sensitive to differences even when normal and anomalous data having similar mean values. The reconstruction probability in VAEs accounts for dissimilarity and variability, allowing nuanced sensitivity to reconstruction based on variable variance \cite{an2015variational}. In this section, we provide more comprehensive details about various VAE architectures mainly focused in this work, including the traditional VAE (with convolutional layers), the VAE with Gaussian Random Field (VAE-GRF) \cite{gangloff2023variational}, and our implemented ViT-VAE.}
\subsubsection{Traditional Variational Autoencoder}
\nhat{A VAE is a generative model specifically designed for likelihood estimation, comprising an encoder and a decoder to capture intricate data distributions\cite{kingma2013auto}. The encoder, denoted as $q_{\phi}(z|x)$, transforms input data $x$ into a latent representation $z$, characterized by parameters $\mu_{\phi}(x)$ and $\sigma_{\phi}(x)$ following a Gaussian distribution. The reparameterization trick is employed to efficiently sample the latent representation, aiding in both training and the generation of diverse samples.

The decoder, represented as $p_{\theta}(\hat{x}|z)$, utilizes the sampled latent representation $z$ to reconstruct the input data $\hat{x}$. The model is trained to minimize the reconstruction error, measured by the negative log-likelihood of the generated data given the latent representation.

VAE incorporates variational inference to estimate the posterior distribution $q_{\phi}(z|x)$. This is achieved by minimizing the Kullback-Leibler (KL) divergence between the learned distribution and a predefined distribution, usually a standard normal distribution $p(z) = \mathcal{N}(0, I)$. The KL divergence term encourages the model to learn a latent space adhering to the desired distribution. The overall objective function includes a reconstruction term, measuring the negative log-likelihood of generated data, and a KL divergence term penalizing the deviation from the predefined distribution.

From the literature, VAE has demonstrated strengths in achieving meaningful latent space representations. These representations facilitate the generation of diverse and realistic samples during the inference phase. The model's ability to efficiently capture complex data distributions and its success in likelihood estimation contribute to its significance in various applications. The training process involves minimizing the total loss, which is the average of individual losses over the training dataset, enabling VAE to effectively learn and generalize from the data. For a deeper insights into the formulation of VAEs, the reparameterization trick, and the objective function, please refer to the original paper of VAE}

\nhat{\textbf{Remark:} A variational autoencoder (VAE) represents an improved version of an autoencoder, integrating regularization methods to counter overfitting and establish favorable characteristics in the latent space for efficient generative capabilities. Operating as a generative model, VAEs pursue a comparable goal to generative adversarial networks.}
\subsubsection{Variational Autoencoder with Gaussian Random Field prior}
\begin{figure*}
  \centering
  \includegraphics[width=0.85\textwidth]{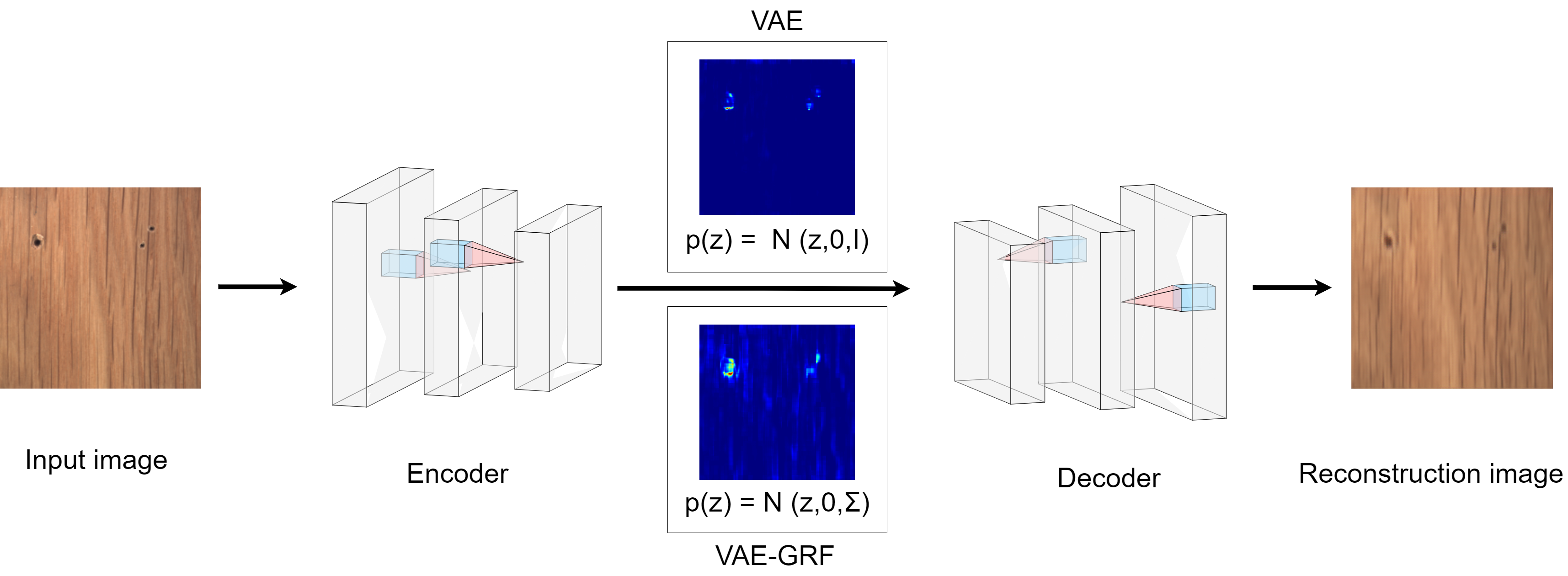} 
  \caption{Illustration of the differences between VAE and VAE-GRF architectures as well as their anomal map differencies \cite{gangloff2023unsupervised}
  .}
  \label{fig:VAE architecture}
\end{figure*}

\nhat{In the VAE-GRF model \cite{gangloff2023unsupervised}, notable strengths have been identified without delving into specific formula details. The VAE-GRF distinguishes itself by maintaining an equivalent number of parameters in the encoder, latent space, and decoder as the traditional VAE, eliminating the need for supplementary modules. Importantly, this model introduces only two additional scalar parameters, the range and variance of the prior, to enhance outcomes with minimal computational expense. The refinement in prior modeling is reported to contribute to improved results, showcasing the model's efficiency and effectiveness, in particular for texture images data.

The VAE-GRF model features a stochastic encoding network that maps input data to a convolutional latent space associated with realizations of a random variable. The introduction of a zero-mean stationary and toroidal Gaussian Random Field as the latent space prior is a key innovation. This GRF prior, with factorized dimensions, brings about enhanced efficiency and computational parallelizability. The model extends the classical VAE by incorporating GRFs, presenting a strict generalization that has been reported to improve performance in various applications. Figure \ref{fig:VAE architecture} illustrates the main difference between the standard VAE and VAE-GRF, lying on their feature distributions within the latent space.

\textbf{Remark: }VAE-GRF represents a robust model with distinct improvements and strengths. Its refined modeling approach, relevance in AD tasks for textured images, and efficiency without added computational costs collectively position VAE-GRF as a promising alternative to the traditional VAE baseline. The model's versatility further underscores its potential for widespread adoption across various domains and applications.
}

\subsubsection{Vision-Transformer Variational Autoencoder}
\nhat{The Vision Transformer (ViT) \cite{dosovitskiy2020image} is a novel architecture for image classification tasks that departs from traditional convolutional neural network (CNN) designs by leveraging self-attention mechanisms. In this paper, we explore the ViT to extract the latent representation of the image. This involves extracting the output from the intermediate layer of the ViT model. The latent representation can be considered as the feature representation of the image captured by the model before passing through the decoder for reconstruction.

The fundamental idea behind ViT is to treat an image as a sequence of fixed-size non-overlapping patches, which are then linearly embedded into vectors\cite{dosovitskiy2020image}. These patches serve as the input tokens for a Transformer architecture, originally designed for natural language processing tasks. In ViT, the image is divided into a grid of patches, and each patch is linearly embedded into a vector. These patch embedding are flattened and serve as the input to the Transformer encoder. The architecture of the ViT-VAE model can be visualized in Figure \ref{fig:vit-vae}.

\begin{figure*}
  \centering
  \includegraphics[width=0.85\textwidth]{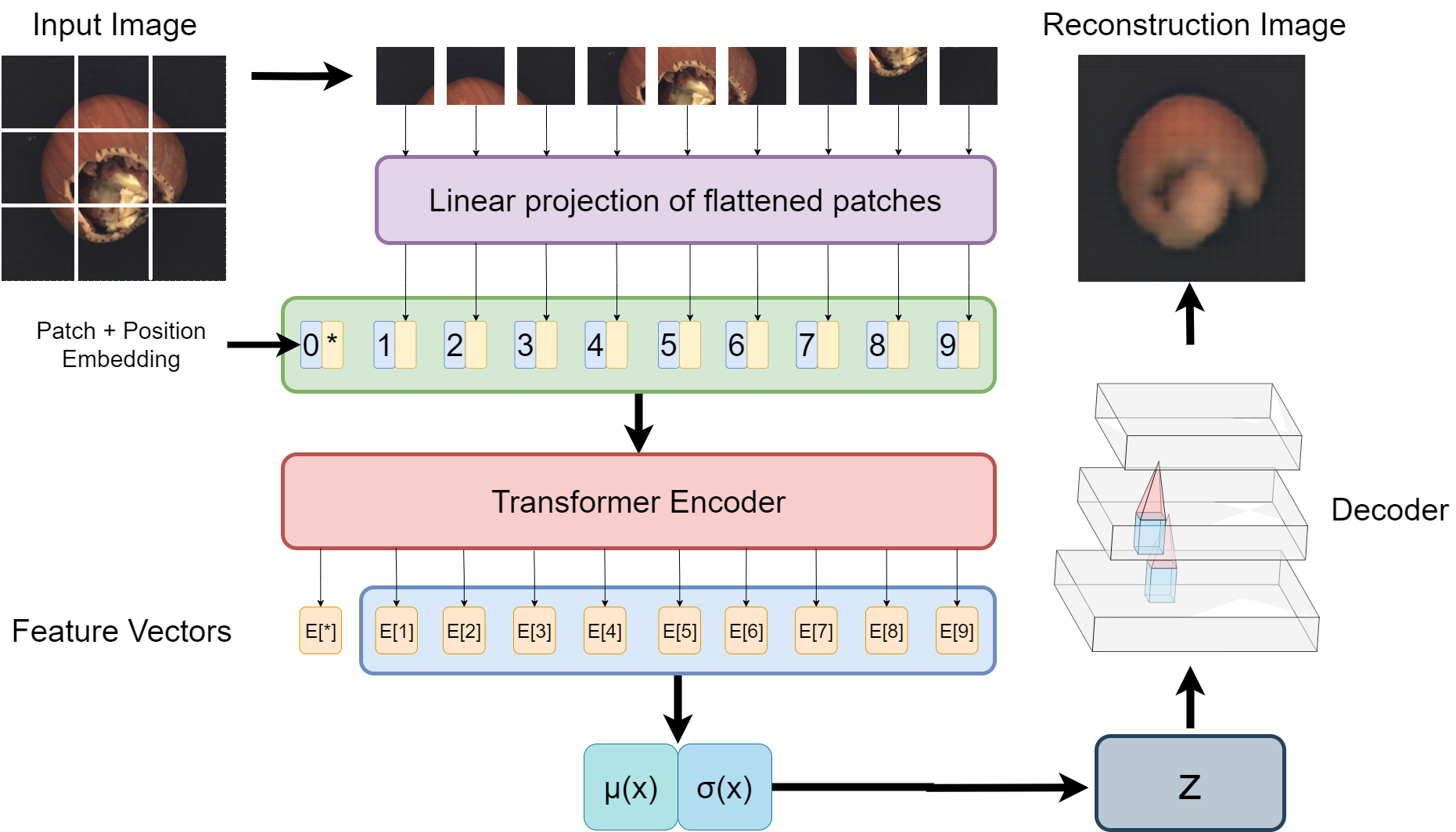}
  \caption{Architecture of ViT-VAE model \cite{lee2022anovit}}
  \label{fig:vit-vae}
\end{figure*}

As observed from the figure, the resulting sequence of embedded vectors is fed into
the Transformer encoder, which consists of multiple layers of
self-attention and feed-forward sub-layers. The self-attention
mechanism allows the model to capture relationships between
different patches in the image. The output of the Transformer encoder serves as a powerful representation of the original image and is employed for downstream tasks. Notably, previous research has highlighted ViT's exceptional performance in basic image processing problems, showcasing its versatility and effectiveness in tasks such as image classification. The ViT architecture's departure from traditional CNN designs, treating images as sequences of patches and leveraging self-attention mechanisms, has led to breakthroughs in image understanding.

\textbf{Remark: } In summary, the VAE-ViT model's strengths lie in its utilization of the ViT architecture, demonstrating success in basic image processing problems. The model efficiently processes images by dividing them into patches, leveraging self-attention mechanisms, and showcasing the versatility of the Vision Transformer in extracting meaningful features for downstream tasks.
}

\subsection{Evaluation Metrics}

\hoang{






We leverage the AD metric proposed in \cite{gangloff2023unsupervised} which is the combination of two metrics including the SSM (Structure-based similarity measure from the reconstruction images) and MAD (alignment map from the latent space). Readers are invited to \cite{gangloff2023unsupervised} 
 for more information. We note that, such metrics can also be used to mark anomaly at pixel level. We evaluate the model using ROCAUC of these values.
}

\section{Experiments}

\subsection{Setup}
\tung{Our experiments were conducted on a machine of 8-GB RAM with NVIDIA GeForce RTX 3050 4GB GPU. The implementation was done using Python with Pytorch 2.0 framework. For hyper-parameter setting, we used 100 epochs and set the batch size to 8, and the learning rate to $10^{-4}$. We reduced the batch size and training epochs from \cite{gangloff2023unsupervised} due to hardware limitation. 

For VAE, we used ResNet18 backbone and set the z dimension of the model to 256 with an feature map size of 32, and the $\beta$ correlation to 1. The same setup was used for VAE-GRF but with other model specialized hyperparameters, including correlation type of the model (we use identity correlation for this experiment), as done in \cite{gangloff2023unsupervised}. However, it should be noted that the settings were done differently for ViT-VAE as ViT does not work the same as CNN (in the standard VAE). ViT requires accurate patch size to capture the image feature in which will also directly affect the latent dimension setting of VAE. In particular, we set the latent dimension of ViT VAE to 384 and latent image size to 14, which is based on the prior implementation of \cite{lee2022anovit}. 

}
\subsection{Datasets}


\trung{To perform our comparative study, two public datasets were used for benchmarking, including the MVTec AD \cite{bergmann2019mvtec} and the MIAD dataset \cite{bao2023miad}. Each of them is now described.}



\subsubsection{The MVTec dataset \cite{bergmann2019mvtec}}
\trung{The MVTec AD dataset  has gained widespread recognition as a benchmark for anomaly detection in industrial scenarios. The dataset comprises 5354 high-resolution images, with 3629 allocated for training and 1725 for testing purposes. The 15 distinct classes within the dataset encompass various industrial objects and scenes \cite{bergmann2019mvtec}


The training set exclusively includes images without any defects, establishing a baseline for normalcy in an industrial context. In contrast, the test set encompasses both defect-free images and those containing various types of defects. These anomalies range from surface defects on objects to structural anomalies, such as distorted object parts, and defects manifested by the absence of specific object components. Notably, the defects present in the test images are manually generated and labeled with pixel-precise masks outlining the defective areas.

Nevertheless, this dataset exhibits two weaknesses. The first issue pertains to the limited number of images for each class, with the class having the highest number of training images totaling 381, a quantity considered insufficient. Another problem arises from the overuse of the MVTec dataset, resulting in the overutilization of current research focused on MVTec. This underscores the necessity for a more recent dataset specifically designed for anomaly detection, that we now describe.



.


}

\subsubsection{The MIAD dataset \cite{bao2023miad}}
\nhat{This dataset is a novel dataset for unsupervised anomaly detection in various maintenance inspections\cite{bao2023miad}. The dataset contains more than 10,000 high-resolution color images in seven outdoor industrial scenarios, such as photovoltaic modules, wind turbine blades, and metal welding. The images are generated by 3D graphics software and cover both surface and logical anomalies with pixel-precise ground truth.

\begin{figure}[!ht]
    \centering
    \includegraphics[width=0.9\linewidth]{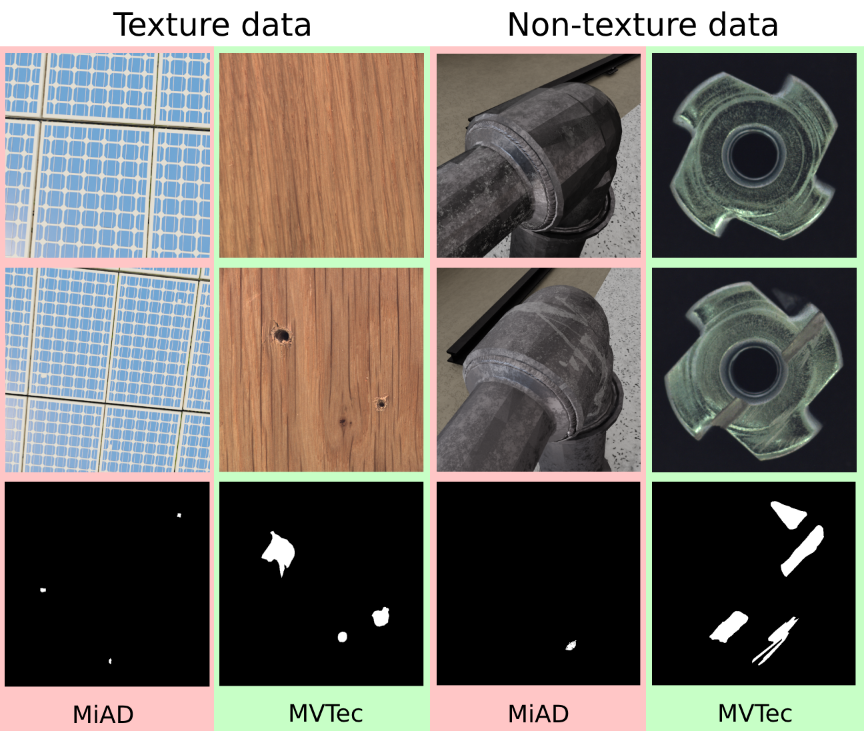}
    \caption{Examples of good and anomalous samples of from the MVTec and the MiAD datasets.}
    \label{fig:example}
\end{figure}


Similar to the MVTec dataset, the MiAD dataset is organized into training and test sets. The training set comprises anomaly-free images, while the test set includes both anomalous and anomaly-free images. In contrast to MVTec, MiAD boasts a significantly larger number of training images, specifically 10,000 images for each class, and a minimum of 1,200 testing images. This notable augmentation stems from the synthetic nature of the MiAD dataset. Notably, MiAD places emphasis on outdoor perspectives captured in uncontrolled environments, incorporating diverse camera viewpoints, complex backgrounds, and object surface degradation. This distinctive focus introduces a new and challenging image domain, contrasting with models predominantly tailored for the MVTec dataset.

Despite MiAD encompassing seven classes, our study restricts its focus to four surface anomaly classes: Electrical Insulator, Metal Welding, Photovoltaic Module, and Wind Turbine. The remaining three classes pertain to logical anomaly detection, a facet beyond the scope of the present investigation. In Figure \ref{fig:example}, we shows some good and anomalous image samples within texture and non-texture materials from the two datasets. We observe that these contain anomaly regions that are quite different in numbers, sizes and forms.

}


\subsection{Comparative results}
\hoang{

In this section, our primary focus revolves around a comprehensive performance comparison of VAE, VAE-GRF and ViT-VAE. It is crucial to acknowledge that GRF prior is specifically tailored for texture data. Therefore, we separate the dataset into two sub-classes, namely texture and non-texture data.

Table~\ref{table:table 1} highlight the difference in performance of the three models across the 15 classes of the MVTec dataset. There are 6 classes classified as texture which are carpet, grid, leather, tile, wood, and hazelnut, and 9 classes as non-texture, namely bottle, cable, capsule, metal nut, pill, screw, toothbrush, transistor, and zipper. Overall, the ViT-VAE model shows significant dominance in performance in non-texture data while, other models may be competitive for some texture images. It is noted that VAE-GRF does not provide good performance in comparison with other models in texture data where it is expected to have good result, according to \cite{gangloff2023unsupervised}.

Table~\ref{table:table 2} shows the record from the experiments with MiAD dataset. In our experiment, the three logical classes from the entire dataset was not included since our focus was on surface (structure) anomalous images. These include Electrical Insulator, Metal Welding, Photovoltaic Module and Wind Turbine. The result on this dataset shown a more competitive result of the three models, with a slightly better performance yielded by VAE-GRF. More discussions will be provided in the next section.
}

\begin{table}[ht]
\centering

\begin{tabular}{lccc}
\toprule
\multirow{2}{*}{Category} & \multicolumn{3}{c}{MAD * SSM} \\
&  VAE & VAE-GRF & ViT-VAE  \\
\midrule
\multicolumn{4}{l}{\textbf{Texture}} \\ 
carpet &  0.88 $\pm$ 0.13 & 0.86 $\pm$ 0.11 & \textbf{\underline{0.92 $\pm$ 0.10}} \\
grid & 0.89 $\pm$ 0.10 & 0.80 $\pm$ 0.12 & \textbf{\underline{0.94 $\pm$ 0.05}} \\
leather &  0.84 $\pm$ 0.16 & \textbf{\underline{0.90 $\pm$ 0.12}} & 0.72 $\pm$ 0.24 \\
tile &  \textbf{\underline{0.84 $\pm$ 0.10}} & 0.68 $\pm$ 0.17 & 0.80 $\pm$ 0.14 \\
wood & \textbf{\underline{0.80 $\pm$ 0.20}} & 0.73 $\pm$ 0.20 & 0.68 $\pm$ 0.19 \\
hazelnut & 0.97 $\pm$ 0.02 & 0.97 $\pm$ 0.02 & \textbf{\underline{0.98 $\pm$ 0.01}} \\
\textbf{Average} & 0.87 $\pm$ 0.11 & 0.82 $\pm$ 0.12  & \textbf{\underline{0.88 $\pm$ 0.12}} \\
\midrule
\multicolumn{4}{l}{\textbf{Non-Texture}} \\
bottle  & 0.80 $\pm$ 0.11 & 0.81 $\pm$ 0.11 & \textbf{\underline{0.85 $\pm$ 0.09}} \\
cable & 0.63 $\pm$ 0.19 & 0.60 $\pm$ 0.24 & \textbf{\underline{0.67 $\pm$ 0.17}} \\
capsule & 0.84 $\pm$ 0.17 & 0.87 $\pm$ 0.15 & \textbf{\underline{0.92 $\pm$ 0.13}} \\
metal nut &0.76 $\pm$ 0.09 & 0.80 $\pm$ 0.09 & \textbf{\underline{0.82 $\pm$ 0.07}} \\
pill& 0.84 $\pm$ 0.16 & 0.87 $\pm$ 0.12 & \textbf{\underline{0.87 $\pm$ 0.17}} \\
screw & 0.92 $\pm$ 0.06 & 0.88 $\pm$ 0.10 & \textbf{\underline{0.94 $\pm$ 0.04}} \\
toothbrush & 0.84 $\pm$ 0.10 & 0.84 $\pm$ 0.07 & \textbf{\underline{0.87 $\pm$ 0.06}} \\
transistor  & 0.69 $\pm$ 0.15 & 0.67 $\pm$ 0.14 & \textbf{\underline{0.72 $\pm$ 0.17}} \\
zipper & 0.86 $\pm$ 0.10 & 0.82 $\pm$ 0.11 & \textbf{\underline{0.89 $\pm$ 0.11}} \\
\textbf{Average}  & 0.79 $\pm$ 0.12  & 0.79 $\pm$ 0.12& \textbf{\underline{0.84 $\pm$ 0.11}} \\
\bottomrule
\end{tabular}
\caption{Comparative results yielded by the three models on the MVTec dataset.}
\label{table:table 1}
\end{table}



\begin{table}[ht]
\centering

\begin{tabular}{lcccc}
\toprule
\multirow{3}{*}{Anomaly} & \multirow{3}{*}{Category} & \multicolumn{3}{c}{MAD * SSM} \\
& & VAE & VAE-GRF & ViT-VAE\\
\midrule
\multirow{4}{*}{Structure}& Electical Insulator  & \textbf{\underline{0.87}} & 0.84 & 0.86 \\
& Metal Welding  & 0.69 & \textbf{\underline{0.82}} & 0.71 \\
& Photovoltaic Module & \textbf{\underline{0.98}} & 0.93 & 0.97 \\
& Wind Turbine  & 0.92 & 0.93 & \textbf{\underline{0.93}} \\
\midrule
\textbf{Average} &  & 0.87 & \textbf{\underline{0.88}} & 0.87 \\
\bottomrule
\end{tabular}
\caption{Comparative results yielded by the three models on the MiAD dataset.}
\label{table:table 2}
\end{table}

\begin{figure*}[t]
  \centering
  \includegraphics[width=0.99\textwidth]{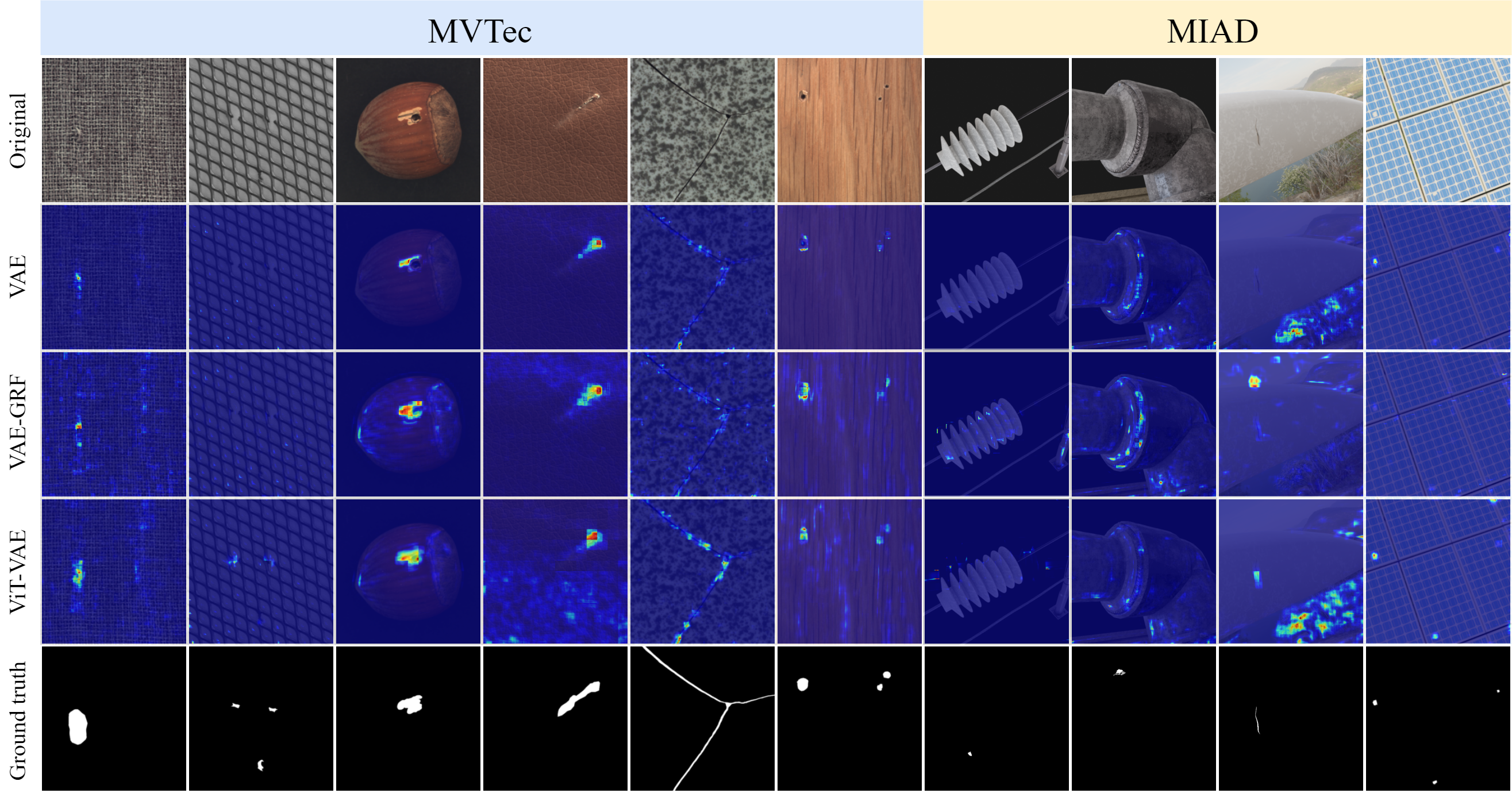}
  \caption{Illustrations of anomaly maps provided by the three models for some MVTec and MIAD samples.}
  \label{fig:miad and mvtec exp}
\end{figure*}

\subsection{Discussion}
\hoang{
In this section, we go in-dept into understand the experiment result and underline the important aspects of the three comparative models. 

\subsubsection{MVTec experiments}
\hoang{
The aim of our experiments on this dataset is to verify the performance of our implemented ViT-VAE compared to the VAE and VAE-GRF whose results were reported in \cite{gangloff2023unsupervised} (but not for all 15 classes). 
From the experimental results showed Table \ref{table:table 1}, there are two important remarks that can be drawed:
\begin{itemize}
\item First, the performance of VAE-GRF on the MVTec dataset is significantly lower than the record in the original paper \cite{gangloff2023unsupervised}, hence we suspect that this is due to the gap in the number of training epochs and the difference in hyperparameter settings. In \cite{gangloff2023unsupervised}, the paper claimed that VAE-GRF requires detail hypothesis and settings to maintain a tracable model. Indeed, to achieve the performance in the original paper in wood and tile, VAE-GRF requires extra hyperparameter tuning in its Matern correlation and $\beta$ in $\beta$ELBO. However, to ensure the equality comparison, we keep the hyperparameter settings of VAE-GRF the same for each classes in this paper, thus leading to the low performance of VAE-GRF. This suggests that to use VAE-GRF, the information about data characteristics should be taken into consideration to find an optimal hyperparameter setting.

\item Secondly, ViT-VAE shows excellent performance in most of the MVTec's classes. By leveraging the ViT architecture, ViT-VAE exhibits an enhanced ability to capture long-range dependencies with reduced inductive bias, albeit at the expense of necessitating a larger volume of data for optimal performance, as asserted by Thisanke et al. \cite{thisanke2023semantic}. Remarkably, ViT-VAE manifests commendable results, particularly excelling in hazelnut, screw, and carpet categories— the top three classes in terms of dataset volume. Conversely, instances where ViT-VAE exhibits comparatively lower performance coincide with classes characterized by a limited volume of data.
\end{itemize}
}

\subsubsection{MiAD experiments}
\hoang{
The MiAD experiment focuses on the robustness of VAE-based anomaly detection methods. To achieve this, we follow the approach of the MiAD dataset, which involves incorporating variations in viewpoint and background into both the training and testing data \cite{bao2023miad}. Upon examining the results of the experiment (showed in Table \ref{table:table 2}), two interesting points can be emerged.

\begin{itemize}
\item First, in the conducted experiments (some illustrations are showed in Figure \ref{fig:miad and mvtec exp}), both VAE and ViT-VAE encounter challenges in handling random backgrounds from the Wind Turbine dataset. Conversely, VAE-GRF does not encounter this issue but still mislabels the anomaly positions. This discrepancy can be elucidated by the fundamental principle of the VAE-GRF prior, wherein a more stationary configuration is related to the improvement of model performance, and vice versa. Specifically, the differences in viewpoints of the training data show a significant effect on the reconstruction output of VAE-GRF. In conclusion, none of the VAE models in this study shows an advantage on the MiAD dataset, thus indicating that MiAD is a challenging dataset for future research.

\item Secondly, the MiAD dataset can also be used as a robustness test for AD frameworks, as demonstrated in our research. By not training the model on the full training dataset, the inconsistency between training and testing data would help researchers clearly identify a model that can still function despite domain shifts. This demonstrates that the MiAD dataset holds immense potential and should be utilized more in the future, along with the classical MVTec dataset.
\end{itemize}
}
}

\section{Conclusion}
\hoang{
Despite its status as the earliest generative algorithm, the Variational Autoencoder (VAE) continues to maintain prominence within the niche of anomaly detection. In pursuit of a comprehensive understanding, we conducted experiments to assess the performance of VAE models in this specific task and clarify their crucial characteristics. Notably, VAE-GRF necessitates meticulous hyperparameter tuning to yield favorable outcomes, whereas the Vision Transformer-based VAE (ViT-VAE) demonstrates robust performance even with a limited number of training epochs.

Furthermore, our observations indicate that the MiAD dataset presents itself as a promising resource. Distinguished by variations in both image domain and anomalies, MiAD holds significant potential for future research endeavors and warrants more extensive uses in anomaly detection studies in future research works.

}



\bibliographystyle{ieeetr}
\bibliography{ref}

\end{document}